\documentclass[letterpaper, 10 pt, conference]{ieeeconf}

\IEEEoverridecommandlockouts  

\usepackage{graphics} 
\usepackage{epsfig} 
\usepackage{mathptmx} 
\usepackage{times} 
\usepackage{amsmath} 
\usepackage{amssymb}  
\usepackage{color}
\usepackage[linesnumbered,ruled,vlined]{algorithm2e}
 \usepackage{url}
 \usepackage{hyperref}
 \usepackage[all]{hypcap}

\newcommand{\urlwofont}[1]
{
\urlstyle{same}\url{#1}
}

\newtheorem{definition}{Definition}
\newtheorem{thm}{Theorem}

\newtheorem{prop}{Proposition}
\newtheorem{problem}{Problem}

\newtheorem{examp}{Example}
\SetAlgoVlined

\title{\LARGE \bf Control of Noisy Differential-Drive Vehicles from Time-Bounded Temporal Logic Specifications }

\author{Igor Cizelj and Calin Belta
\thanks{This work was partially supported by 
the ONR  MURI under grant N00014-10-10952
and by the NSF under grant CNS-0834260. }
\thanks{The authors are with the Division of Systems Engineering at Boston University, Boston, MA 02215, USA. Email:
        {\tt\small $\{$icizelj,cbelta$\}$@bu.edu}.}%
}

\begin{document}
\maketitle
\thispagestyle{empty}
\pagestyle{empty}

\begin{abstract}
We address the problem of controlling a noisy differential drive mobile robot such that the probability of satisfying a specification given as a Bounded Linear Temporal Logic (BLTL) formula over a set of properties at the regions in the environment is maximized. We assume that the vehicle can determine its precise initial position in a known map of the environment. However, inspired by practical limitations, we assume that the vehicle is equipped with noisy actuators and, during its motion in the environment, it can only measure the angular velocity of its wheels using limited accuracy incremental encoders. 
Assuming the duration of the motion is finite, we map the measurements to a Markov Decision Process (MDP).  
We use recent results in Statistical Model Checking (SMC) to obtain an MDP control policy that maximizes the probability of satisfaction. 
We translate this policy to a vehicle feedback control strategy and show that the probability that the vehicle satisfies the specification in the environment is bounded from below by the probability of satisfying the specification on the MDP.
We illustrate our method with simulations and experimental results. 
\end{abstract}

\section{Introduction}
Robot motion planning and control has been widely studied in the last twenty years. In ``classical'' motion planning problems \cite{Lavalle:planning}, the specifications are usually restricted to simple primitives of the type ``go from $A$ to $B$ and avoid obstacles'', where $A$ and $B$ are two regions of interest in some environment. Recently, temporal logics, such as Linear Temporal Logic (LTL) and Computational Tree Logic (CTL) (\cite{baier:principles,Clarke1999}) have become increasingly popular for specifying robotic tasks (see, for example  \cite{1428622}, \cite{kress-gazit:whereswaldo?}, \cite{karaman:vehicle}, \cite{kloetzer:fully}, \cite{Murray2009}, \cite{6016581}). It has been shown that temporal logics can serve as rich languages capable of specifying complex motion missions such as ``go to region $A$ and avoid region $B$ unless regions $C$ or $D$ are visited''.

In order to use existing model checking tools for motion planning (see \cite{baier:principles}), many of the above-mentioned works 
rely on the assumption that the motion of the vehicle in the environment can be modeled as a finite system \cite{Clarke1999} that is either deterministic (applying an available action triggers a unique transition \cite{DingRecedingHorizon}) or nondeterministic (applying an available action can enable multiple transitions, with no information on their likelihoods \cite{KlBe-HSCC08-book}). Recent results show that, if sensor and actuator noise models can be obtained from empirical measurements or an accurate simulator, then the robot motion can be modeled as a Markov Decision Process (MDP), and probabilistic temporal logics, such as Probabilistic CTL (PCTL) and Probabilistic LTL (PLTL), can be used for motion planning and control (see \cite{LWAB10}).

However, robot dynamics are normally described by control systems with state and control variables evaluated over infinite domains. A widely used approach for temporal logic verification and control of such a system is through the construction of a finite abstraction (\cite{tabuada:lineartime,Girard:approximately, kloetzer:fully,YoBe-TAC-2011}).
Even though recent works discuss the construction of abstractions for stochastic systems \cite{julius:approximations, Abate:Markov,D'Innocenzo:Approximate}, the existing methods are either not applicable to robot dynamics or are computationally infeasible given the size of the problem in most robotic applications. 

In this paper, we consider a vehicle whose performance is measured by the completion of time constrained temporal logic tasks. In particular, we provide a conservative solution to the problem of controlling a stochastic differential drive mobile robot such that the probability of satisfying a specification given as a Bounded Linear Temporal Logic (BLTL) formula over a set of properties at the regions in the environment is maximized. Motivated by a realistic scenario of an indoor vehicle leaving its charging station, we assume that the vehicle can determine its precise initial position in a known map of the environment. The actuator noise is modeled as a random variable with a continuous probability distribution supported on a bounded interval, where the distribution is obtained through experimental trials. Also, we assume that the vehicle is equipped with two limited accuracy incremental encoders, each measuring the angular velocity of one of the wheels, as the only means of measurement available. 

Assuming the duration of the motion is finite, through discretization, we map the incremental encoder measurements to an MDP. By relating the MDP to the vehicle motion in the environment, the vehicle control problem becomes equivalent to the problem of finding a control policy for an MDP such that the probability of satisfying the BLTL formula is maximized. Due to the size of the MDP, finding the exact solution is prohibitively expensive. We trade-off correctness for scalability and we use computationally efficient techniques based on sampling. Specifically, we use recent results in Statistical Model Checking for MDPs (\cite{StatisticalModelCheckingforMDPs}) to obtain an MDP control policy and a Bayesian Interval Estimation (BIE) algorithm (\cite{Zuliani:2010:BSM:1755952.1755987}) to estimate the probability of satisfying the specification. We show that the probability that the vehicle satisfies the specification in the original environment is bounded from below by the maximum probability of satisfying the specification on the MDP under the obtained control policy. 

The main contribution of this work lies in bridging the gap between low level sensory inputs and high level temporal logic specifications. We develop a framework for the synthesis of a vehicle feedback control strategy from such specifications based on a realistic model of an incremental encoder. This paper extends our previous work (\cite{ProbSafeControlofNoisyDubinsVehicle}) of controlling a stochastic version of Dubins vehicle such that the probability of satisfying a temporal logic statement, given as a PCTL formula, over some environmental properties, is maximized. Specifically, the approach presented here allows for richer temporal logic specifications, where the vehicle performance is measured by the completion of time constrained temporal logic tasks. Additionally, in order to deal with the increase in the size of the problem we use computationally efficient techniques based on sampling. In \cite{StatisticalModelCheckingforMDPs}, the authors use Statistical Model Checking for MDPs to solve a motion planning problem for a vehicle moving on a finite grid and knowing its state precisely, at all times, when the task is given as a BLTL formula.  We adopt this approach to control a vehicle with continuous dynamics and allowing for uncertainty in its state. 

The remainder of the paper is organized as follows. In Sec. \ref{sec:preliminaries}, we introduce the necessary notation and review some preliminaries. We formulate the problem and outline the approach in Sec. \ref{sec:problem_formulation_and_approach}.
In Sec. \ref{sec:generating_a_trace} - \ref{sec:generating_a_trace_under_the_position_uncertainty} we explain the construction of the MDP and the relation between the MDP and the motion of the vehicle in the environment. The vehicle control policy is obtained in Sec. \ref{sec:vehicle_control_strategy}. Case studies and experimental results illustrating our approach are presented in Sec. \ref{sec:case_study}. We conclude with final remarks and directions for future work in Sec. \ref{sec:conclusion}. 

\label{sec:intro}
\section{Preliminaries}
\label{sec:preliminaries}

In this section we provide a short and informal introduction to Markov Decision Processes (MDP) and Bounded Linear Temporal Logic (BLTL). For details about MDPs the reader is referred to \cite{baier:principles} and for more information about BLTL to \cite{DBLP:conf/cmsb/JhaCLLPZ09} and \cite{Zuliani:2010:BSM:1755952.1755987}. 
\begin{definition}[MDP]
\label{def:MDP}
A Markov Decision Process (MPD) is a tuple $M=(S,s_0,Act,A, P)$, where 
$S$ is a finite set of states; $s_0 \in S$ is the initial state;  $Act$ is a finite set of actions; $A: S\rightarrow 2^{Act}$ is a function specifying the enabled actions at a state $s$; $P: S \times Act \times S \rightarrow [0,1]$ is a transition probability function such that for all states $s \in S$ and actions $a \in A(s)$: $\sum_{s' \in S}P (s,a,s')=1$, and for all actions $a \notin A(s)$ and $s'\in S$: $P (s,a,s')=0$; 
\end{definition}

A control policy for an MDP resolves nondeterminism in each state $s$ by providing a distribution over the set of actions enabled in $s$.
\begin{definition}[MDP Control Policy]
\label{def:mdp_control_policy}
A control policy $\mu$ of an MDP $M$ is a function $\mu(s,a): S \times Act \rightarrow [0,1]$, s.t., $\sum_{a \in A(s)}\mu(s,a)=1$ and $\mu(s,a)>0$ only if $a$ is enabled in $s$. A control policy for which either $\mu(s,a)=1$ or $\mu(s,a)=0$ for all pairs $(s,a) \in S \times Act$ is called deterministic. 
\end{definition}

We employ Bounded Linear Temporal Logic (BLTL) to describe high level motion specifications. BLTL is a variant of Linear Temporal Logic (LTL) (\cite{baier:principles}) which requires only paths of bounded size. A detailed description of the syntax and semantics of BLTL is beyond the scope of this paper and can be found in \cite{DBLP:conf/cmsb/JhaCLLPZ09} and \cite{Zuliani:2010:BSM:1755952.1755987}. Roughly, formulas of BLTL  are constructed by connecting properties from a set of proposition $\Pi$ using Boolean operators ($\neg$ (negation), $\wedge$ (conjunction), $\vee$ (disjunction)), and temporal operators ($\mathbf{U}^{\leq t}$ (bounded until), $\mathbf{F}^{\leq t}$ (bounded finally), and $\mathbf{G}^{\leq t}$ (bounded globally), where $t \in \mathbb{R}^{\geq0}$ is the time bound parameter). The semantics of BLTL formulas are given over infinite traces $\sigma=(o_1,t_1)(o_2,t_2)\ldots$, $o_i \in 2^{\Pi}$, $t_i \in \mathbb{R}^{\geq 0}$, $i \geq1$, where $o_i$ is the set of satisfied propositions and $t_i$ is the time spent satisfying $o_i$. A trace satisfies a BLTL formula $\phi$ if $\phi$ is true at the first position of the trace; $\mathbf{F}^{\leq t}\phi_1$ means that $\phi_1$ will be true within $t$ time units; $\mathbf{G}^{\leq t}\phi_1$ means that $\phi_1$ will remain true for the next $t$ time units; and $\phi_1 \mathbf{U}^{\leq t} \phi_2$ means that $\phi_2$ will be true within the next $t$ time units  and $\phi_1$ remains true until then. More expressivity can be achieved by combining the above temporal and Boolean operators.

\section{Problem Formulation and Approach}
\label{sec:problem_formulation_and_approach}
\subsection{Problem Formulation}
\label{sec:proble_formulation}
A differential drive mobile robot (\cite{Lavalle:planning}) is a vehicle having two main wheels, each of which is attached to its own motor, and a third wheel which passively rolls along preventing the robot from falling over. In this paper, we consider a stochastic version of a differential drive mobile robot, which captures actuator noise: 
\begin{equation}
\begin{bmatrix}
 \dot x\\  \dot y\\ \dot \theta
\end{bmatrix}
= 
\begin{bmatrix}
\frac{r}{2} (u_r+\epsilon_{r}+u_l+\epsilon_l)\cos(\theta)\\ 
\frac{r}{2} (u_r+\epsilon_{r}+u_l+\epsilon_l)\sin(\theta)\\ 
\frac{r}{L}(u_r+\epsilon_{r}-u_l-\epsilon_l)
\end{bmatrix}, \text{ } u_r  \in U_r, \text{ } u_l \in U_l,
\label{eq:eq1}
\end{equation}
where $(x,y) \in \mathbb{R}^2$ and $\theta \in [0,2 \pi)$ are the position and orientation of the vehicle in a world frame,  $u_r$ and $u_l$ the control inputs (angular velocities before being corrupted by noise), $U_r$ and $U_l$ are control constraint sets, and $\epsilon_r$ and $\epsilon_l$ are random variables modeling the actuator noise with continuous probability density functions supported on the bounded intervals $[\epsilon_r^{min},\epsilon_r^{max}]$ and  $[\epsilon_l^{min},\epsilon_l^{max}]$, respectively. $L$ is the distance between the two wheels and $r$ is the wheel radius. 
We denote the state of the system by $q=[x,y,\theta]^T \in SE(2)$. 

Motivated by the fact that the time optimal trajectories for the bounded velocity differential drive robots are composed only of turns in place and straight lines (\cite{Balkcom_2000_3169}), we assume $U_r$ and $U_l$ are finite, but we make no assumptions on the optimality. 
We define 
\begin{align*}
W_i &=\{u+\epsilon | u \in U_i, \epsilon \in [\epsilon_i^{min},\epsilon_i^{max}] \}, \text{ } i \in \{r,l\},
\end{align*}
as the sets of applied control inputs, i.e, the sets of angular wheel velocities that are applied to the system in the presence of noise. We assume that time is uniformly discretized (partitioned) into stages (intervals) of length $\Delta t$, where stage $k$ is from $(k-1)\Delta t$ to $k \Delta t$. The duration of the motion is finite and it is denoted $K \Delta t$ (later in this section we explain how $K$ is determined). We denote the control inputs and the applied control inputs at stage $k$ as $u_i^k \in U_i$, $i \in \{r,l\}$, and $w_i^k \in W_i$, $i \in \{r,l\}$, respectively.

We assume that the vehicle is equipped with two incre- mental encoders, each measuring the applied control input (i.e., the angular velocity corrupted by noise) of one of the wheels. Motivated by the fact that the angular velocity is considered constant inside the given observation stage ([PTPZ07]), the applied controls are considered piecewise constant, i.e., $w_i:[(k-1)\Delta t,k\Delta t] \rightarrow W_i$, $i \in \{r,l\}$, are constant over each stage. 

{\emph{Incremental encoder model:}}
As shown in \cite{4510607}, the measurement resolution of an incremental encoder is constant and for encoder $i$ we denote it as $\Delta \epsilon_i$, $i \in \{r,l\}$. Given $\Delta \epsilon_i$ and $[\epsilon^{min}_i,\epsilon^{max}_i]$, $i \in \{r,l\}$, then the following holds: $\exists n_i \in \mathbb{Z}^{+}$ s.t. $n_i \Delta \epsilon_i=|\epsilon^{max}_i-\epsilon^{min}_i|$, $i \in \{r,l\}$.
For more details see Sec. IX where we also explain how to obtain the measurement resolutions and the probability density functions. Then, $[\epsilon_i^{min},\epsilon_i^{max}]$ can be partitioned\footnote[1]{Throughout the paper, we relax the notion of partition by allowing the endpoints of the intervals to overlap.} into $n_i$ noise intervals of length $\Delta \epsilon_i$: $[\underline{\epsilon}_i^{j_i},\overline{\epsilon}_i^{j_i}]$, $j_i=1,\ldots,n_i$, $i \in \{r,l\}$. We denote the set of all noise intervals $\mathcal{E}_i=\{[\underline{\epsilon}_i^{1},\overline{\epsilon}_i^{1}],\ldots,[\underline{\epsilon}_i^{n_i},\overline{\epsilon}_i^{n_i}]\}$, $i \in \{r,l\}$. At stage $k$, if the applied control input is $u_i^k+\epsilon_i$, the incremental encoder $i$ will return measured interval $$[\underline{w}_i^k,\overline{w}_i^k]=[u_i^k+\underline{\epsilon}_i,u_i^k+\overline{\epsilon}_i],$$ where $\epsilon_i \in [\underline{\epsilon}_i, \overline{\epsilon}_i] \in \mathcal{E}_i$, $i \in \{r,l\}$. In Fig. 1 we give an example. The pair of measured intervals at stage $k$, $([\underline{w}_r^k,\overline{w}_r^k],[\underline{w}_l^k,\overline{w}_l^k])$, returned by the incremental encoders, is denoted $\mathbb{W}^k$. 
\begin{figure}[htb]
\begin{center}
\includegraphics[width=0.4\textwidth]{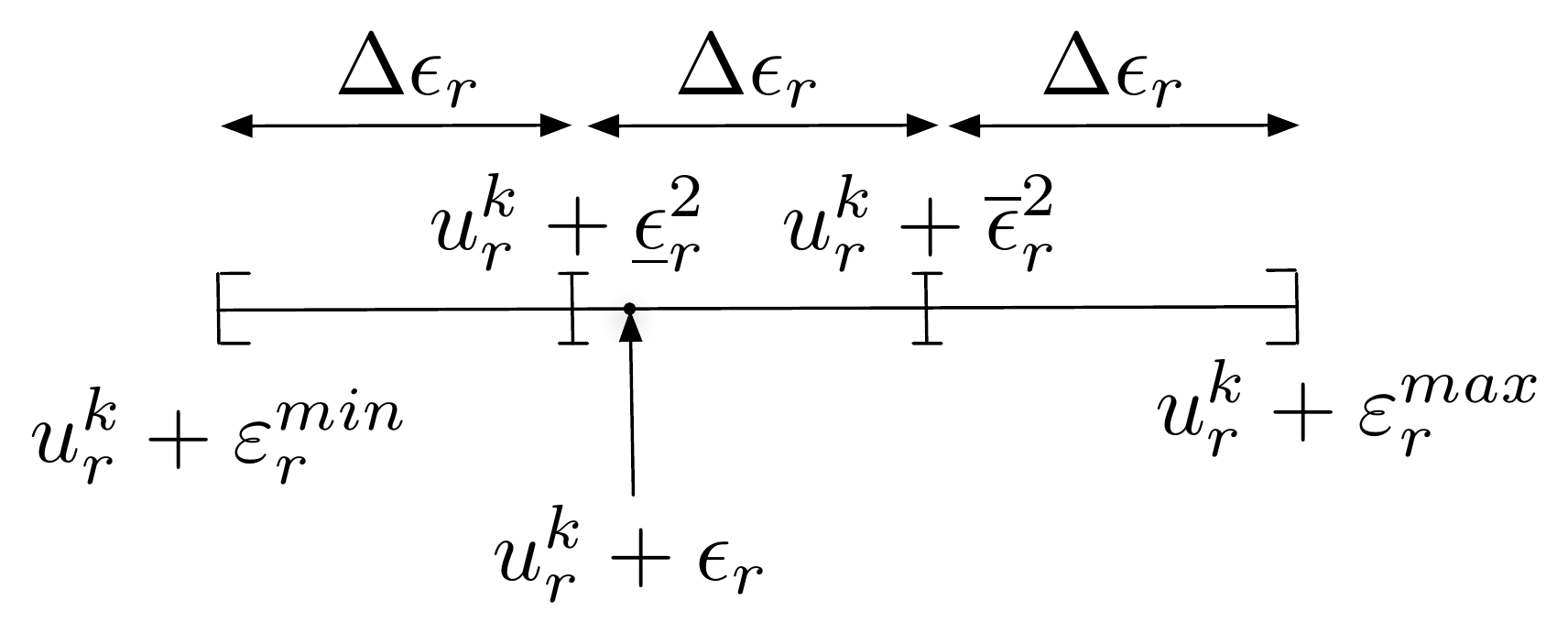}
\label{figure:fig1}
\end{center}
\caption{Let $n_r=3$, i.e, $[\epsilon_r^{min},\epsilon_r^{max}]$ is partitioned into $3$ noise intervals of length $\Delta \epsilon_r$, $\mathcal{E}_r=\{[\underline{\epsilon}_r^1,\overline{\epsilon}_r^1],[\underline{\epsilon}_r^2,\overline{\epsilon}_r^2],[\underline{\epsilon}_r^3,\overline{\epsilon}_r^3]\}$. Assume the applied control input at stage $k$ is $u_r^k+\epsilon_r$, such that $\epsilon_r \in [\underline{\epsilon}_r^2,\overline{\epsilon}_r^2]$. Then, the incremental encoder $r$, at stage $k$, will return measured interval $[\underline{w}_r^k,\overline{w}_r^k]=[u_r^k+\underline{\epsilon}_r^2,u_r^k+\overline{\epsilon}_r^2]$.}
\end{figure}

The vehicle moves in a planar environment in which a set of non-overlapping regions of interest, denoted $R$, is present. Let $\Pi$ 
be the set of propositions satisfied at the regions in the environment.  One of these propositions, denoted by $\pi_u \in \Pi$, signifies that the corresponding regions are {\ttfamily{unsafe}}.
In this work, the motion specification is expressed as a BLTL formula $\phi$ over $\Pi$:
\begin{equation}
\label{eq:eq2}
\phi = \neg \pi_{u} \mathbf{U}^{\leq T_1} ( \varphi_1 \wedge \neg \pi_u \mathbf{U}^{\leq T_2} ( \varphi_2 \wedge  \ldots  \wedge \neg \pi_u \mathbf{U}^{\leq T_f}\varphi_f ) ),\end{equation} 
$f \in \mathbb{Z}^+$, and $\varphi_j$, $\forall j \in \{1,\ldots,f\}$, is of the following form:
\begin{equation*}
\varphi_j=\mathbf{G}^{\leq \tau_j^1}(\bigvee_{\pi \in \Pi_j^1}\pi) \vee \ldots \vee \mathbf{G}^{\leq \tau_j^{n_j}}(\bigvee_{\pi \in \Pi_j^{n_j}}\pi),
\end{equation*}
where $n_j\in \mathbb{Z}^+$, $\forall_{n=1,\ldots,n_j}\Pi_j^n \subset \Pi \setminus \pi_{u}$, $\forall_{n=1,\ldots,n_j}\tau_j^n \in \mathbb{R}^{\geq0}$ and $T_j \in \mathbb{R}^{\geq0}$. 

\begin{examp}
\label{examp:examp1}
Consider the environment shown in Fig. \ref{fig:fig2}. Let $\Pi=\{\pi_u,\pi_p,\pi_t,\pi_d\}$,  where $\pi_u,\pi_p,\pi_t,\pi_d$ label the {\ttfamily{unsafe}}, {\ttfamily{pick-up}}, {\ttfamily{test}} and the {\ttfamily{drop-off}} regions, respectively. Let the motion specification be as follows:

{\emph{
Start from an initial state $q_{init}$ and reach a {\ttfamily{pick-up}} region within $T_1$ time units to pick up a load. After entering the {\ttfamily{pick-up}} region reach a {\ttfamily{test}} region within $T_2$ time units and stay in it at least $\tau_2$ time units. Finally, after entering the {\ttfamily{test}} region reach a {\ttfamily{drop-off}} region within $T_3$ time units to drop off the load. Always avoid the {\ttfamily{unsafe}} regions. 
}}

The specification translates to BLTL formula $\phi$:
\begin{equation}
\label{eq:eq3}
\phi=\neg \pi_u \mathbf{U}^{\leq T_1}(\pi_p \wedge \neg \pi_u \mathbf{U}^{\leq T_2}(\mathbf{G}^{\tau_2} \pi_t \wedge \neg \pi_u \mathbf{U}^{\leq T_3} \pi_d)).
\end{equation}

\begin{figure}[htb]
\begin{center}
\includegraphics[width=0.5\textwidth]{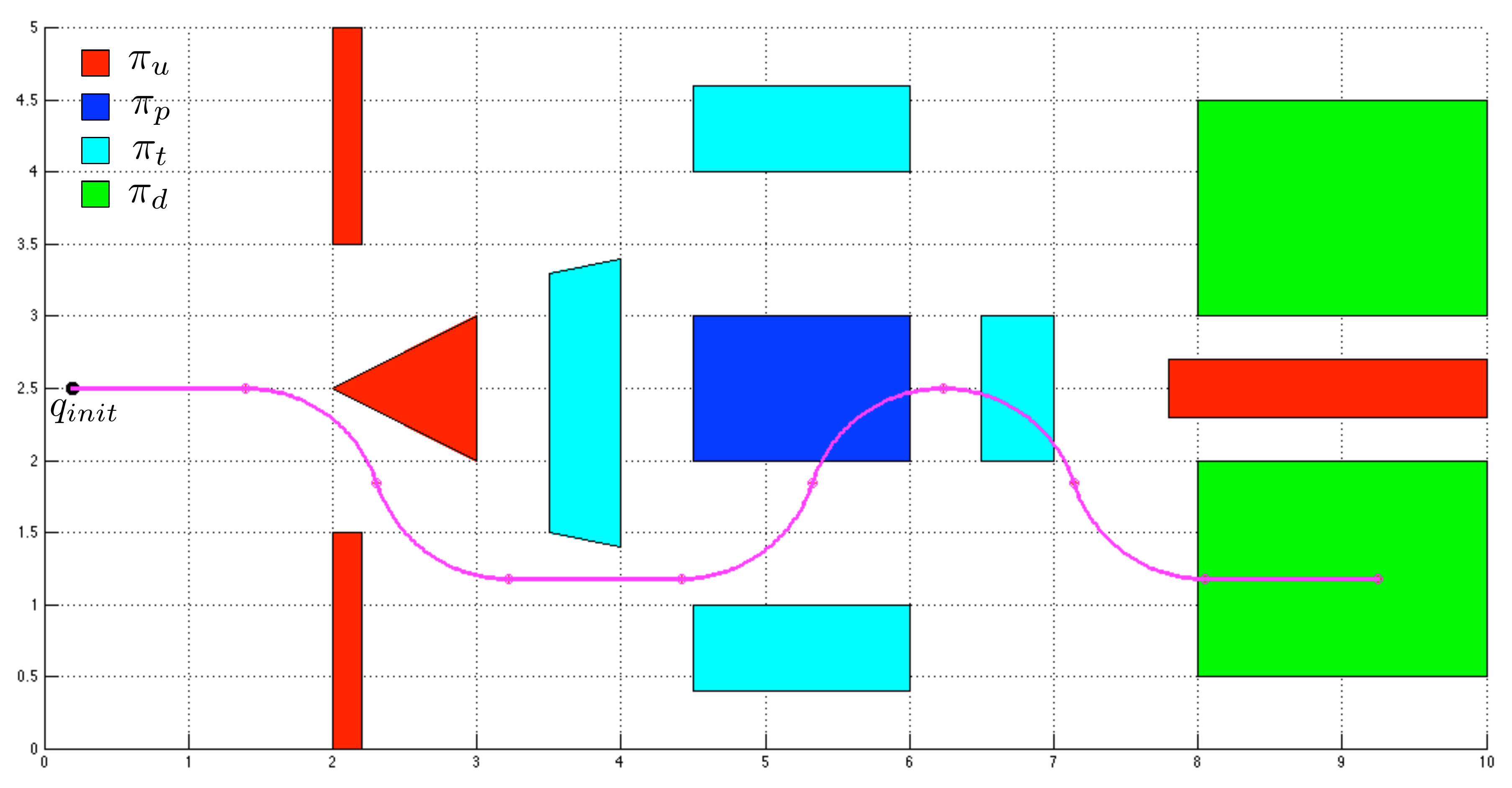}
\end{center}
\caption{An example environment with the regions of interest. The {\ttfamily{unsafe}}, {\ttfamily{pick-up}}, {\ttfamily{test}} and the {\ttfamily{drop-off}} regions are shown in red, blue, cyan and green, respectively. A sample state (position) trajectory of the system is shown in magenta.}
\label{fig:fig2}
\end{figure}
\end{examp}

We assume that the vehicle can precisely  determine its initial state $q_{init}=[x_{init},y_{init},\theta_{init}]$, in a known map of the environment. While the vehicle moves, incremental encoder measurements $\mathbb{W}^k$ are available at each stage $k$. We define a {\emph{vehicle control strategy}} as a map that takes as input a sequence of pairs of measured intervals $\mathbb{W}^1 \mathbb{W}^2 \ldots \mathbb{W}^{k-1}$, and returns control inputs $u_r^k \in U_r$ and $u_l^k \in U_l$ at stage $k$. We are ready to formulate the main problem we consider in this paper:

\begin{problem}
\label{problem:main}
Given a set of regions of interest $R$ satisfying propositions from a set $\Pi$, a vehicle model described by Eqn. (\ref{eq:eq1}) with initial state $q_{init}$, a motion specification expressed as a BLTL formula $\phi$ over $\Pi$ (Eqn. (\ref{eq:eq2})), find a vehicle control strategy that maximizes the probability of satisfying the specification.
\end{problem}

To fully specify Problem \ref{problem:main}, we need to define the satisfaction of a BLTL formula $\phi$ by a trajectory $q:[0,K\Delta t] \rightarrow SE(2)$ of the system from Eqn. (\ref{eq:eq1}). Formal definition is given in Sec. \ref{sec:generating_a_trace}. Informally, $q(t)$ produces a finite trace $\sigma=(o_1,t_1)(o_2,t_2)\ldots(o_l,t_l)$, $o_i \in \Pi \cup \emptyset$, $t_i \in \mathbb{R}^{\geq 0}$, $i \geq1$, where $o_i$ is the satisfied proposition\footnote[2]{Since the regions of interest are non-overlapping it follows that $o_i \in \Pi \cup \emptyset$.}
and $t_i$ is the time spent satisfying $o_i$, as time evolves. A trajectory $q(t)$ satisfies BLTL formula $\phi$ if and only if the generated trace satisfies the formula. Given $\phi$, for the duration of the motion we use the smallest $K \in \mathbb{Z}^{+}$ for which model checking a trace is well defined, i.e., the smallest $K$ for which the maximum nested sum of time bounds  (see \cite{Zuliani:2010:BSM:1755952.1755987}) is at most $K \Delta t$.

\subsection{Approach}
\label{sec:approach}
In this paper, we develop a suboptimal solution to Problem \ref{problem:main} consisting of three steps. First, we define a finite state MDP that captures every sequence realization of pairs of measurements returned by the incremental encoders. States of the MDP correspond to the sequences of pairs of measured intervals and the actions correspond to the control inputs. 

Second, we find a control policy for the MDP that maximizes the probability of satisfying BLTL formula $\phi$. Because of the size of the MDP, finding the exact solution is computationally too expensive.  We decided to trade-off correctness for scalability and we use computationally efficient technique based on system sampling. We use recent results in SMC for MDPs (\cite{StatisticalModelCheckingforMDPs}) to obtain an MDP control policy and BIE algorithm (\cite{Zuliani:2010:BSM:1755952.1755987}) to estimate the probability of satisfying $\phi$.

Finally, since each state of the MDP corresponds to a unique sequence of pairs of measured intervals, we translate the control policy to a vehicle control strategy. In addition, we show that the probability of satisfying $\phi$, in the original environment, is bounded from below by the probability of satisfying the specification on the MDP under the obtained control policy. 

\section{Generating a trace}
\label{sec:generating_a_trace}
In this section we explain how, given a state trajectory the corresponding trace is generated.
Let us denote $[\pi]=\{(x,y) \in \mathbb{R}^2|(x,y) \in \cup_{r \in R_{\pi}}r\}$ as the set of positions that satisfy proposition $\pi$, where $R_{\pi} \subseteq R$ is the set of regions labeled with proposition $\pi$.

\begin{definition}[Generating a trace] The {\emph{trace corresponding to a state trajectory}} $q(t)=[x(t),y(t),\theta(t)]^T$ is a  finite sequence $\sigma=(o_1,t_1)(o_2,t_2)\ldots(o_l,t_l)$, $o_i \in \Pi \cup \emptyset$, $t_i \in [0,K \Delta t]$, $i=1,\ldots,l$, $l \geq1$, where $o_i$ is the satisfied proposition and $t_i$ is the time spent satisfying $o_i$, generated according to the following rules, for all $t,t',\tau \in [0,K \Delta t]$:\\
$\bullet$ $o_1=\pi \in \Pi$ {\bf{iff}} $(x(0),y(0)) \in [\pi]$ and $o_1=\emptyset$ otherwise. \\
$\bullet$ Let $o_i$ be the satisfied proposition at some $t$. Then:
\begin{enumerate}
\item {\bf{If}} $o_i=\emptyset$, {\bf{then}} $o_{i+1}=\pi \in \Pi$, {\bf{iff}} (i) $\exists t'>t$ s.t. $(x(t'),y(t')) \in [\pi]$, and (ii) $\nexists \tau \in [t,'t]$ s.t. $(x(\tau),y(\tau)) \in [\pi']$, $\forall \pi' \in \Pi$  {\bf{and}} $t_{i}=\text{min}_{t \in [\sum_{j=0}^{i-1}t_j,K\Delta t]}\{t| (x(t),y(t)) \in [\pi]\}-\sum_{j=0}^{i-1}t_j$, with $t_0=0$.
\item {\bf{If}} $o_i=\pi \in \Pi$, {\bf{then}} $o_{i+1}=\emptyset$ {\bf{iff}} $\exists t'>t$ s.t. $(x(t'),y(t')) \notin [\pi]$, {\bf{and}} $t_{i}=\text{min}_{t \in [\sum_{j=0}^{i-1}t_j,K\Delta t]}\{t| (x(t),y(t)) \notin [\pi]\}-\sum_{j=0}^{i-1}t_j$, with $t_0=0$.
\end{enumerate}
$\bullet$ Let for $K \Delta t$, $o_l$ be the current satisfied propositions. Then, $t_l=K \Delta t - \sum_{j=1}^{l-1}t_j$.
\label{def:def3}
\end{definition}

A trajectory $q(t)$ satisfies BLTL formula $\phi$ (Eqn. (\ref{eq:eq2})) if and only if the trace generated according to the rules stated above satisfies the formula. Note that, since the duration of the motion is finite, the generated trace is also finite. In \cite{Zuliani:2010:BSM:1755952.1755987} the authors show that BLTL requires only traces of bounded lengths. The fact that the trace $\sigma$ satisfies $\phi$ is denoted $\sigma \vDash \phi$. Given a trace $\sigma$, the $i$-th state of $\sigma$, denoted $\sigma_i$, is $(o_i,t_i)$, $i=1,\ldots,l$. We denote $\sigma |_i$ as the finite subsequence of $\sigma$ that starts in $\sigma_i$. Finally, given a formula $\phi$, we denote subformula $\neg \pi_u \mathbf{U}^{T_j} \varphi_j$ as $\phi_j$, $j = 1, \ldots, f$. 
Using the BLTL semantics one can derive the following conditions to determine whether $\sigma \vDash \phi$:
\begin{definition}[Satisfaction conditions]
Given a trace $\sigma$ and a BLTL formula $\phi$ (Eqn. (\ref{eq:eq2})), let for $j \in \{1,\ldots,f\}$, $i_j, k_j \in \mathbb{N}$ be such that for some $n \in \{1,\ldots, n_j\}$ the following holds:
\begin{enumerate}
\item $o_{i_j+k_j} \in \Pi_j^n$,
\item for each $i_j \leq i <i_j+k_j$, $o_{i} \neq \pi_u$,
\item $\sum_{i=i_j}^{i_j+k_j-1}t_{i} \leq T_j$, and
\item $t_{{i_j}+k_j} \geq \tau_j^n$.
\end{enumerate}
Then, $\sigma |_{i_j} \vDash \phi_j$. If $\forall j \in \{1,\ldots,f\}$, $\exists i_j, k_j \in \mathbb{N}$ s.t. $\sigma|_{i_j} \vDash \phi_j$ where $i_{j+1}=i_j+k_j$ with $i_1=1$, then $\sigma \vDash \phi$.
\label{def:def4}
\end{definition}

\begin{examp}
\label{examp:examp2} 
Consider the environment and the sample state (position) trajectory shown in Fig. $2$. Let $\phi$ be as in Eq. (\ref{eq:eq3}) with the following numerical values for the time bounds: $T_1=6.2$, $T_2 = 2.3$, $\tau_2=0.2$, and $T_3=2.3$. The trajectory generates trace $\sigma = (\emptyset,6.12)(\pi_p,0.75)(\emptyset,0.44)(\pi_t,0.61)(\emptyset,1.66)(\pi_d,1.22)$. The following holds: $\sigma|_1 \vDash \phi_1$ since for $i_1=1$ and $k_1=1$, $o_{2} \in \{\pi_p\}$, $o_1 \neq \pi_u$, $t_1 \leq T_1$; $\sigma|_2 \vDash \phi_2$ since for $i_2=2$ and $k_2=2$, $o_{4} \in \{\pi_t\}$, $o_2,o_3 \neq \pi_u$, $t_2+t_3 \leq T_2$ and $t_4 \geq \tau_2$; and $\sigma|_4 \vDash \phi_3$ since for $i_3=4$ and $k_3=2$, $o_{6} \in \{\pi_d\}$, $o_4,o_5 \neq \pi_u$ and $t_4+t_5 \leq T_3$; Thus, $\sigma \vDash \phi$.
\end{examp}

\section{Construction of an MDP Model}
\label{sec:construction_of_an_mdp_model}
Recall that $\epsilon_i$ is a random variable with a continuous probability density function
supported on the bounded interval $[\epsilon_i^{min},\epsilon_i^{max}]$, $i \in \{r,l\}$.  The probability density functions are obtained through experimental trials (see Sec. IX) and they are defined as follows:
\begin{equation}
\label{eq:eq4}
\text{Pr}(\epsilon_i \in [\underline{\epsilon}_i^{j_i},\overline{\epsilon}_i^{j_i}])=p_i^{j_i},
\end{equation}
$[\underline{\epsilon}_i^{j_i},\overline{\epsilon}_i^{j_i}] \in \mathcal{E}_i$, $j_i=1,\ldots,n_i$, s.t. $\sum_{j_i=1}^{n_i}p_i^{j_i}=1$, $i \in \{r,l\}$. 

An MDP $M$ that captures every sequence realization of pairs of measurements returned by the incremental encoders is defined as a tuple $(S,s_0,Act,A,P)$, where:
\begin{itemize}
\item $S=\cup_{k=1,\ldots,K}\{([u_r+\underline{\epsilon}_r,u_r+\overline{\epsilon}_r],[u_l+\underline{\epsilon}_l,u_l+\overline{\epsilon}_l]) | u_r \in U_r, u_l \in U_l, [\underline{\epsilon}_r,\overline{\epsilon}_r] \in \mathcal{E}_r, [\underline{\epsilon}_l,\overline{\epsilon}_l] \in \mathcal{E}_l\}^k$. The meaning of the state is as follows: $(\mathbb{W}^1,\ldots, \mathbb{W}^k) \in S$, means that at stage $i$, $1 \leq i \leq k$, the pair of measured intervals is $\mathbb{W}^i$. 
\item $s_0=\emptyset$ is the initial state.
\item $Act = \{U_r \times U_l\} \cup \varphi$ is the set of actions, where $\varphi$ is a dummy action.
\item $A:S \rightarrow 2^{Act}$ gives the enabled actions at state $s$: if $|s|=K$, i.e., if the termination time is reached, $A(s)=\varphi$, otherwise $A(s)=\{U_r \times U_l\}$.
\item $P: S \times Act \times S \rightarrow [0,1]$ is a transition probability function constructed by the following rules:
\begin{enumerate}
\item If $s=(\mathbb{W}^1,\ldots,\mathbb{W}^k) \in S$ then $P(s,a,s')=p_r^m p_l^n$ iff $s'=(\mathbb{W}^1,\ldots,\mathbb{W}^k,([u_r+\underline{\epsilon}_r^m,u_r+\overline{\epsilon}_r^m],[u_l+\underline{\epsilon}_l^n,u_l+\overline{\epsilon}_l^n])) \in S$ and $a=(u_r,u_l) \in \{U_r \times U_l\}$ where $m=1,\ldots,n_r$, $n=1,\ldots,n_l$ and $k=1, \ldots, K$; 
\item If $|s|=K$ then $P(s,a,s')=1$ iff $a=\varphi$ and $s'=s$; 
\item $P(s,a,s')=0$ otherwise. 
\end{enumerate}
\end{itemize}
Rule 1) defined above follows from the fact that given $u_r^k$ and $u_l^k$ as the control inputs at stage $k$, the pair of measured intervals at stage $k+1$ is $([u_r^k+\underline{\epsilon}_r^m,u_r^k+\overline{\epsilon}_r^m],[u_l^k+\underline{\epsilon}_l^n,u_l^k+\overline{\epsilon}_l^n])$ with probability $p_r^m p_l^n$, since $\text{Pr}(\epsilon_r \in [\underline{\epsilon}_r^m,\overline{\epsilon}_r^m])=p_r^m$ and $\text{Pr}(\epsilon_l \in [\underline{\epsilon}_l^n,\overline{\epsilon}_l^n])=p_r^n$, which follows from Eqn. (\ref{eq:eq4}) (see the MDP fragment in Fig. 3).
Rule 2) states that if the length of $s$ is equal to $K$, i.e.,  if the termination time is reached, then $A(s)=\varphi$ with $P(s,a,s)=1.$

\begin{figure}[htb]
\label{fig:fig3}
\begin{center}
\includegraphics[width=0.5\textwidth]{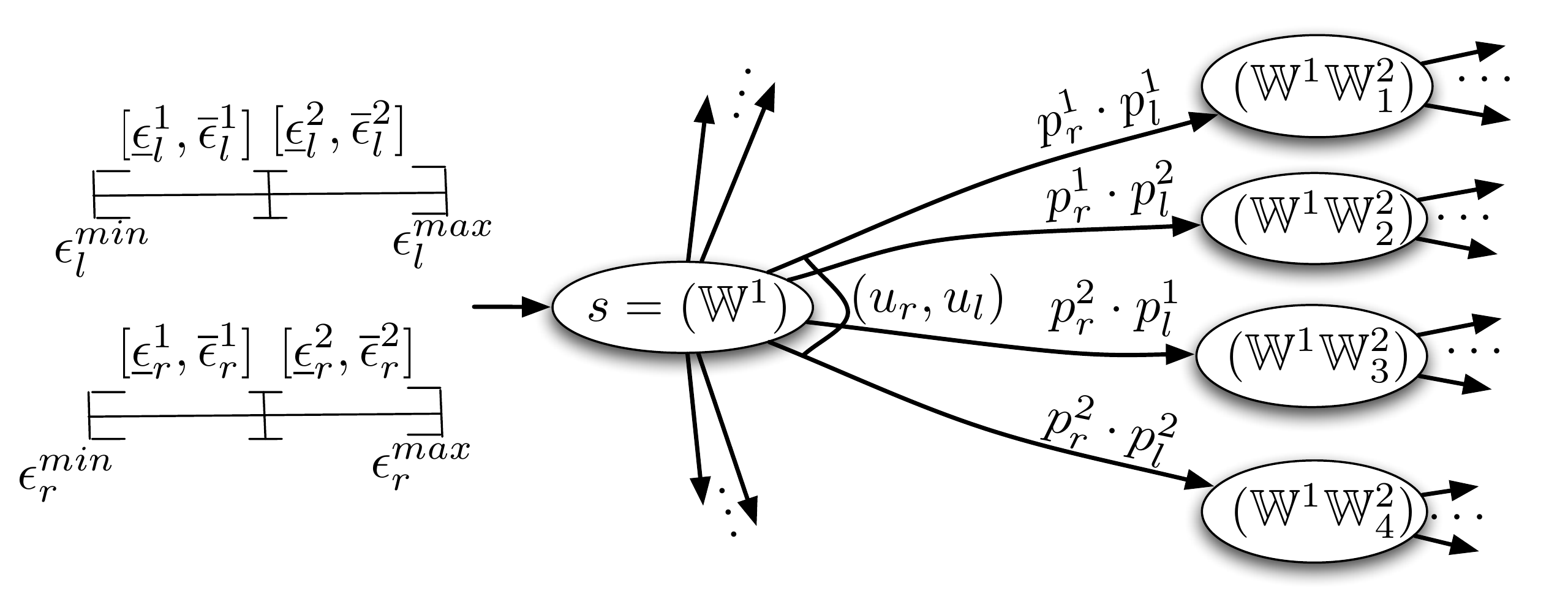}
\caption{A fragment of the MDP $M$ where $n_r=n_l=2$. Thus, $p_r^m=\text{Pr}(\epsilon_r \in [\underline{\epsilon}_r^m,\overline{\epsilon}_r^m])$, for $m=1,2$, and $p_l^n=\text{Pr}(\epsilon_l \in [\underline{\epsilon}_l^n,\overline{\epsilon}_l^n])$, for $n=1,2$. Action $(u_r,u_l) \in A(s)$ enables four transitions. For example, given state $s=(\mathbb{W}^1)$, the new state is $(\mathbb{W}^1\mathbb{W}^2_2)$, where $\mathbb{W}^2_2=([u_r-\underline{\epsilon}_r^1,u_r+\overline{\epsilon}_r^1],[u_l-\underline{\epsilon}_l^2,u_l+\overline{\epsilon}_l^2])$, with probability $p_r^1 \cdot p_l^2$. This corresponds to applied control inputs being equal to $u_r+\epsilon_r$ and $u_l+\epsilon_l$ where $\epsilon_r \in [\underline{\epsilon}_r^1,\overline{\epsilon}_r^1]$ and $\epsilon_l \in [\underline{\epsilon}_l^2,\overline{\epsilon}_l^2]$.}
\end{center}
\end{figure}

\begin{prop}
The model $M$ defined above is a valid MDP, i.e., it satisfies the Markov property and $P$ is a transition probability function.
\end{prop}
{\bf{Proof:}} The proof follows from construction of $P$. Given current state $s \in S$ and an action $a \in A(s)$, the conditional probability distribution of future states depends only on the current state $s$, not on the sequences of events that preceded it (see rule 1) above). Thus, the Markov property holds. In addition, since for every $s$ and $a \in A(s)$: $\sum_{s' \in S}P(s,a,s')=\sum_{m=1}^{n_r}\sum_{n=1}^{n_l}p_r^mp_l^n=\sum_{m=1}^{n_r}p_r^m \sum_{n=1}^{n_l}p_l^n=1$, it follows that $P$ is a valid transition probability function. $\blacksquare$

\section{Position uncertainty}
\label{sec:position_uncertainty}
\subsection{Nominal state trajectory}
\label{sec:nominal_state_trajectory}
For each interval belonging to the set of noise intervals $\mathcal{E}_i$, we define a representative value $\epsilon_i^{j_i}=(\underline{\epsilon}_i^{j_i}+\overline{\epsilon}_i^{j_i})/{2}$, $j_i=1,\ldots,n_i$, $i \in \{r,l\}$, i.e., $\epsilon_i^{j_i}$ is the midpoint of interval $[\underline{\epsilon}_i^{j_i},\overline{\epsilon}_i^{j_i}] \in \mathcal{E}_i$, $i \in \{r,l\}$. We denote the set of representative values as $E_i=\{\epsilon_i^{1},\ldots,\epsilon_i^{n_i}\}$, $i \in \{r,l\}$.

We use $q^k(t)$, $w^k_r$ and $w^k_l$, $t \in [(k-1)\Delta t,k\Delta t]$, $k=1,\ldots,K$, to denote the state trajectory and the constant applied controls at stage $k$, respectively. With a slight abuse of notation, we use $q^k$ to denote the end of state trajectory $q^k(t)$, i.e., $q^k=q^k(k\Delta t)$. Given state $q^{k-1}$, the state trajectory $q^k(t)$ can be derived by integrating the system given by Eqn. (\ref{eq:eq1}) from the initial state $q^{k-1}$, and taking into account the applied controls are constant and equal to $w^k_r$ and $w^k_l$. Throughout the paper, we will also denote this trajectory by $q^k(q^{k-1},w^k_r,w^k_l,t)$, when we want to explicitly capture the initial state $q^{k-1}$ and the constant applied controls $w^k_r$ and $w^k_l$.

Given a path through the MDP: 
\begin{equation}
\label{eq:eq5}
s_0 \xrightarrow{(u_r^1,u_l^1)}s_1\xrightarrow{(u_r^2,u_l^2)}s_2\ldots s_{K-1}\xrightarrow{(u_r^K,u_l^K)}s_K,
\end{equation}
 where $s_k=(\mathbb{W}^1,\ldots,\mathbb{W}^k)$, with $\mathbb{W}^k=([u_r^k+\underline{\epsilon}_r^k,u_r^k+\overline{\epsilon}_r^k],[u_l^k+\underline{\epsilon}_l^k,u_l^k+\overline{\epsilon}_l^k])$, $k=1,\ldots,K$, we define the {\emph{nominal state trajectory}} $q(t)$, $t \in [0,K \Delta t]$, as follows: $$q(t)=q^k(q^{k-1},u_r^k+\epsilon_r^k,u_l^k+\epsilon_l^k,t), \text{ } t \in [(k-1)\Delta t,\Delta t],$$ $k=1,\ldots,K$, where $\epsilon_i^k \in E_i$ is such that $\epsilon_i^k \in [\underline{\epsilon}_i^k,\overline{\epsilon}_i^k]$, $i \in \{r,l\}$ and $q^0=q_{init}$. For every path through the MDP, its nominal state trajectory is well defined. The next step is to define the uncertainty evolution, along the nominal state trajectory, since the applied controls can take any value within the measured intervals.

\subsection{Position uncertainty evolution}
\label{sec:position_uncertainty_evolution}
Since a motion specification is a statement about the propositions satisfied by the regions of interest in the environment, in order to answer whether some state trajectory satisfies BLTL formula $\phi$ it is sufficient to know its projection in $\mathbb{R}^2$. Therefore, we focus only on the position uncertainty.

The position uncertainty of the vehicle when its nominal position is $(x,y) \in \mathbb{R}^2$ is modeled as a disc centered at $(x,y)$ with radius $d \in \mathbb{R}$, where $d$ denotes the distance uncertainty:
\begin{equation}
D((x,y),d)=\{(x'y') \in \mathbb{R}^2 | ||(x,y),(x',y')|| \leq d\},
\label{eq:eq6}
\end{equation}
where $||\cdot||$ denotes the Euclidian distance. Next, we explain how to obtain $d$.

First, let $\Delta \theta \in S^1$ denote the orientation uncertainty. Let $q(t)$, $t \in [0,K \Delta t]$, be the nominal state trajectory corresponding to a path through the MDP (Eqn. (\ref{eq:eq5})). Then, $q(t)$ can be partitioned into $K$ state trajectories: $q^k(t)=q^k(q^{k-1},u_r^k+\epsilon_r^k,u_l^k+\epsilon_l^k,t)$, $t \in [(k-1)\Delta t,\Delta t]$, $k=1,\ldots,K$, where $\epsilon_i^k \in E_i$ is such that $\epsilon_i^k \in [\underline{\epsilon}_i^k,\overline{\epsilon}_i^k] \in \mathcal{E}_i$, $i \in \{r,l\}$ and $q^0=q_{init}$ (see Fig. 4).
 The distance and orientation uncertainty at state $q^k$ are denoted as $d^k$ and $\Delta \theta^k$, respectively. We set $d^k$ and $\Delta \theta^k$ at state $q^k=[x^k,y^k,\theta^k]^T$ equal to: 
\begin{equation}
\begin{array}{l}
\label{eq:eq7}
d^k=\text{max}_{[x',y',\theta']^T \in \mathcal{R}^k} (||(x^k,y^k),(x',y')||)+d^{k-1} \text{ and }\\
\Delta \theta^k=\text{max}_{[x',y',\theta']^T \in \mathcal{R}^k}(|\theta^k-\theta'|),
\end{array}
\end{equation}
where 
\begin{equation}
\label{eq:eq8}
\begin{split}
\mathcal{R}^k=\{q^k([x^{k-1},y^{k-1},\theta^{k-1}+\alpha]^T,u_r^k+\epsilon_r',u_l^k+\epsilon_l',k\Delta t)| \\ \alpha \in \{\Delta \theta^{k-1},-\Delta \theta^{k-1}\}, \epsilon_r' \in \{\underline{\epsilon}_r^k,\overline{\epsilon}_r^k\}, \epsilon_l' \in \{\underline{\epsilon}_l^k,\overline{\epsilon}_l^k\}\},
\end{split}
\end{equation}
for $k=1,\ldots,K$, where $d^0=0$ and $\Delta \theta ^0=0$.

\begin{figure}[htb]
\label{fig:fig3}
\begin{center}
\includegraphics[width=0.49\textwidth]{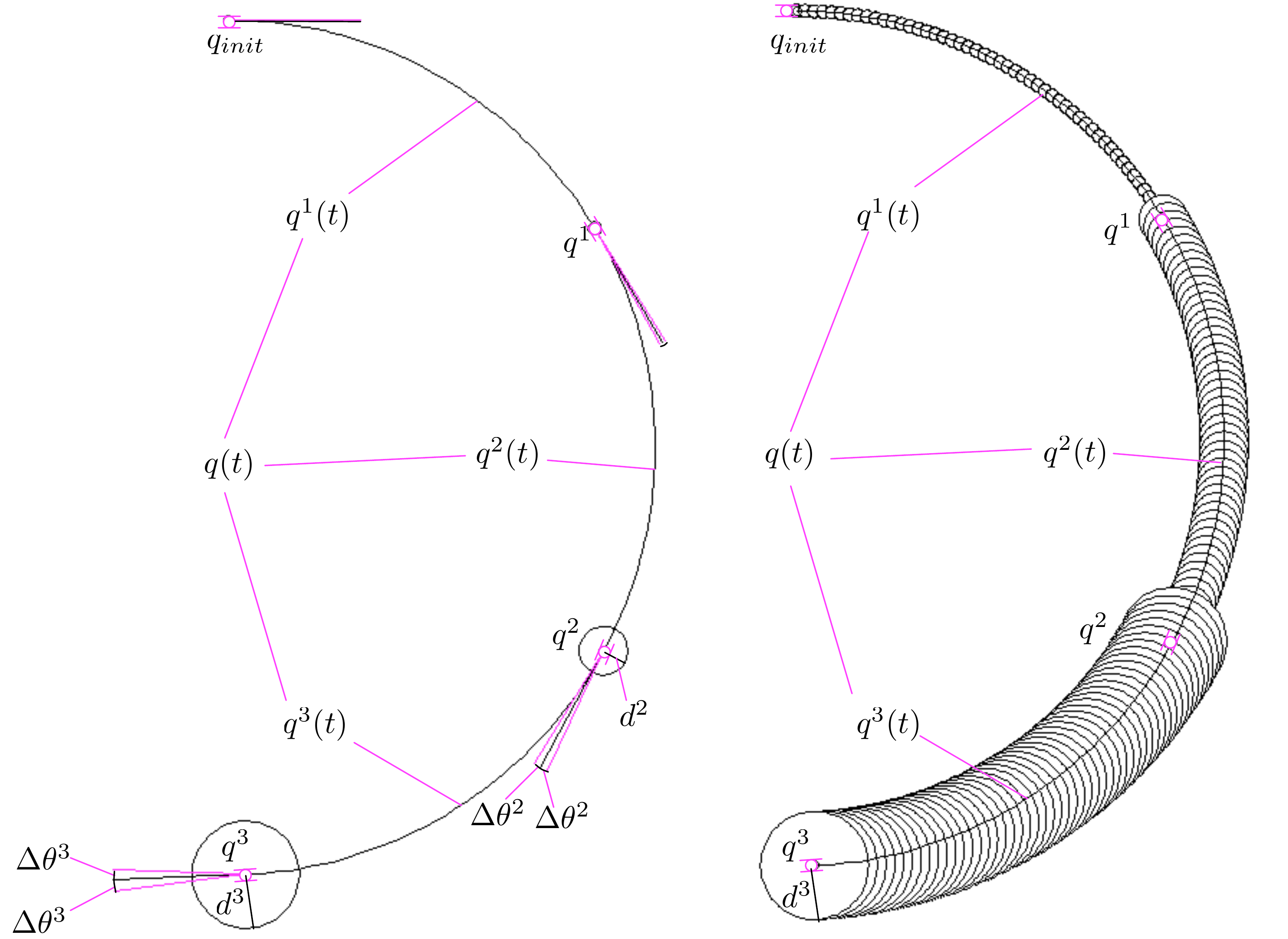}
\caption{Left: Evolution of the position uncertainty along the nominal state trajectory $q(t)=[x(t),y(t),\theta(t)]$, where $q(t)$ is partitioned into $3$ state trajectories, $q^k(t)$, $k=1,2,3$. Right: The conservative approximation of region $D((x(t),y(t)),d(t))$ along $q(t)$, where the distance uncertainty trajectory is $d(t')=d^k(t)$, $t' \in [(k-1)\Delta t, k \Delta  t]$, where $d^k(t)=d^k$, $k=1,2,3$.
}
\end{center}
\end{figure}
Eqn. (\ref{eq:eq7}) and (\ref{eq:eq8}) are obtained using a worst scenario assumption. At stage $k$, the pair of measured intervals is $\mathbb{W}^k=([u_r^k+\underline{\epsilon}_r^k,u_r^k+\overline{\epsilon}_r^k],[u_l^k+\underline{\epsilon}_l^k,u_l^k+\overline{\epsilon}_l^k])$ and we use the endpoints of the measured intervals to define set $\mathcal{R}^k$. $\mathcal{R}^k$ is the smallest set of points in $SE(2)$, at the end of stage $k$, guaranteed to contain (i) the state with the maximum distance (in Euclidian sense) from $q^k$ given that the applied controls at stage $i$ are within the measured intervals at stage $i$, and (ii) the state with the maximum orientation difference compared to $q^k$ given that the applied controls at stage $i$ are within the measured intervals at stage $i$, $i=1,\ldots,k$. (for more details about $\mathcal{R}^k$ see \cite{Fraichard98pathplanning}). An example is given in Fig. 4.

From Eqn. (\ref{eq:eq7}) and (\ref{eq:eq8}) it follows that, given a nominal state trajectory $q(t)$, $t \in [0,K \Delta t]$, the distance uncertainty increases as a function of time. The way it changes along $q(t)$ makes it difficult to characterize the exact shape of the position uncertainty region. Instead, we use a conservative approximation of the region. We define $d:[0,K \Delta t] \rightarrow \mathbb{R}$ as an {\emph{approximate distance uncertainty trajectory}} and we set $d(t)=d^k$, $t \in [(k-1)\Delta t,k\Delta t]$, $k=1,\ldots,K$, i.e., we set the distance uncertainty along the state trajectory $q^k(t)$ equal to the maximum value of the distance uncertainty along $q^k(t)$, which is at state $q^k$. An example illustrating this idea is given in Fig. 4.

\begin{prop}
\label{prop:prop2}
Given a path through the MDP $M$ (Eqn. (5)), and the corresponding $q(t)$ and $d(t)$, $t \in [0,K \Delta t]$, as defined above, then any state trajectory $q'(t)=q^k(q^{k-1},u_r^k+\epsilon_r^{k'},u_l^k+\epsilon_l^{k'},t)$, $t \in [(k-1)\Delta t, k \Delta t]$, $k=1,\ldots,K$, where $q^0=q_{init}$, $\epsilon_r^{k'} \in [\underline{\epsilon}_r^k,\overline{\epsilon}_r^k]$ and $\epsilon_l^{k'} \in [\underline{\epsilon}_l^k,\overline{\epsilon}_l^k]$, is within the uncertainty region, i.e.,  $(x'(t),y'(t)) \in D((x(t),y(t)),d(t))$, $\forall t \in [0,K \Delta t]$.
\end{prop}
{\bf{Proof:}} The proof follows from the definition of the approximate distance uncertainty trajectory and Eqn. (\ref{eq:eq6}), (\ref{eq:eq7}) and (\ref{eq:eq8}). $\blacksquare$

\section{Generating a trace under the position uncertainty}
\label{sec:generating_a_trace_under_the_position_uncertainty}
Let $q(t)$ be a nominal state trajectory with the distance uncertainty trajectory $d(t)$, $t \in [0,K\Delta t]$. In this subsection we introduce a set of conservative rules according to which the trace corresponding to the uncertainty region $D((x(t),y(t)),d(t))$ is generated. This rules guarantee that if the generated trace satisfies $\phi$ (Eqn. (2)) then any state (position) trajectory, inside $D((x(t),y(t)),d(t))$, will satisfy $\phi$. 

\begin{definition}[Generating a trace under uncertainty]
\label{def:def5}
The \emph{trace corresponding to an uncertainty region} $D((x(t),y(t)),d(t))$ is a finite sequence $\sigma=(o_1,t_1)(o_2,t_2)\ldots,(o_l,t_l)$, $o_i \in \Pi \cup \emptyset$, $t_i \in [0,K \Delta t]$, $i=1,\ldots,l$, $l \geq 1$, where $o_i$ is the satisfied proposition and $t_i$ is the time spent satisfying $o_i$, generated according to the following rules, for all $t,t',\tau \in [0,K \Delta t]$:\\
$\bullet$ $o_1= \pi \in \Pi \setminus \pi_u$ {\bf{iff}} $D((x(0),y(0),d(0)) \subseteq [\pi]$, $o_1=\pi_u$ {\bf{iff}} $D((x(0),y(0),d(0)) \cap [\pi_u] \neq \emptyset$ and $o_1=\emptyset$ otherwise.\\
$\bullet$ Let $o_i$ be the satisfied proposition at some $t$. Then:
\begin{enumerate}
\item  {\bf{If}} $o_i=\pi \in \Pi \setminus \pi_u$, {\bf{then}} $o_{i+1}=\emptyset$ {\bf{iff}} $\exists t'>t$ s.t. $D((x(t'),y(t')),d(t')) \not\subseteq [\pi]$ {\bf{and}} $t_{i}=\text{min}_{t \in [\sum_{j=0}^{i-1}t_j,K\Delta t]}\{t| D((x(t),y(t)),d(t)) \not\subseteq [\pi]\}-\sum_{j=0}^{i-1}t_j$, with $t_0=0$.

\item  {\bf{If}} $o_i=\pi_u$, {\bf{then}} $o_{i+1}=\emptyset$ {\bf{iff}} $\exists t'>t$ s.t. $D((x(t'),y(t')),d(t')) \cap [\pi_u] = \emptyset$ {\bf{and}} $t_{i}=\text{min}_{t \in [\sum_{j=0}^{i-1}t_j,K\Delta t]}\{t|  D((x(t),y(t)),d(t))\cap [\pi_u] = \emptyset \}-\sum_{j=0}^{i-1}t_j$, with $t_0=0$.

\item  {\bf{If}} $o_i=\emptyset$, {\bf{then}} $o_{i+1}= \pi \in \Pi \setminus \pi_u$, {\bf{iff}} 
\begin{enumerate}
\item $\exists t'>t$ s.t. $D((x(t'),y(t')),d(t')) \subseteq [\pi]$, 
\item $\nexists \tau \in [t,t']$ s.t. $D((x(\tau),y(\tau)),d(\tau)) \subseteq [\pi']$, $\forall \pi' \in \Pi \setminus \pi_u$
\item $\nexists \tau \in [t,t']$ s.t. $D((x(\tau),y(\tau)),d(\tau)) \cap [\pi_u] \neq \emptyset$ 
\end{enumerate} {\bf{and}} $t_{i}=\text{min}_{t \in [\sum_{j=0}^{i-1}t_j,K\Delta t]}\{t|  D((x(t),y(t)),d(t)) \subseteq [\pi] \}-\sum_{j=0}^{i-1}t_j$, with $t_0=0$.

\item {\bf{If}} $o_i=\emptyset$, {\bf{then}} $o_{i+1}= \pi_u$, {\bf{iff}} 
\begin{enumerate}
\item $\exists t'>t$ s.t. $D((x(t'),y(t')),d(t')) \cap [\pi_u] \neq \emptyset$, 
\item $\nexists \tau \in [t,t']$ s.t. $D((x(\tau),y(\tau)),d(\tau)) \subseteq [\pi']$, $\forall \pi' \in \Pi \setminus \pi_u$, and 
\end{enumerate} {\bf{and}} $t_{i}=\text{min}_{t \in [\sum_{j=0}^{i-1}t_j,K\Delta t]}\{t|  D((x(t),y(t)),d(t)) \cap [\pi_u] \neq \emptyset\}-\sum_{j=0}^{i-1}t_j$, with $t_0=0$.
\end{enumerate}
$\bullet$ Let for $K \Delta t$,  $o_l$ be the current satisfied proposition. Then $t_l=K \Delta t - \sum_{j=1}^{l-1}t_j$.
\end{definition}

In Fig. 5 we show an uncertainty region and the corresponding trace generated according to rules stated above. Next, we show that if the trace corresponding to an uncertainty region satisfies $\phi$, then any state (position) trajectory inside the uncertainty region also satisfies $\phi$.

\begin{prop}
\label{prop:prop3} 
Let $D((x(t),y(t)),d(t))$ be the uncertainty region corresponding to a path through the MDP $M$  (Eqn. (\ref{eq:eq5}))
 and let $q'(t)$ be any state trajectory as defined in Prop. \ref{prop:prop2}. Let $\sigma^D=(o_1^D,t_1^D)\ldots(o_k^D,t_k^D)$ and $\sigma^{q'}=(o_1^{q'},t_1^{q'})\ldots(o_l^{q'},t_l^{q'})$ be the corresponding traces. Given BLTL formula $\phi$ (Eqn. (\ref{eq:eq2})), if $\sigma^D \vDash {\phi}$, then $\sigma^{q'} \vDash {\phi}$.
\end{prop}
{\bf{Proof:}} First, we state two relations between the given traces:
\begin{enumerate} 
\item Let $o_i^D=\pi \in \Pi \setminus \pi_u$ for some $i \in \{1,\ldots,k\}$.
Then, the following holds: $\exists j \in \{1,\ldots,l\}$ such that $o_j^{q'}=\pi$ and $t_i^D \leq t_j^{q'}$. \\
Informally, if $t^D_i$ is the time $D((x(t),y(t)),d(t))$ spent inside the region satisfying proposition $\pi$, then $q'(t)$ will spend at least $t_i^D$ time units inside that region.

\item Let $o_i^D=\pi \in \Pi \setminus \pi_u$ and $o_{i'}^D=\pi' \in \Pi \setminus \pi_u$ for some $i,i' \in \{1,\ldots,k\}$, $i'>i$. Then, the following holds: $\exists j,j' \in \{1,\ldots,l\}$, $j' > j$ such that $o_j^{q'}=\pi$ and $o_{j'}^{q'}=\pi'$. In addition, $\sum_{h=j}^{j'-1}t^{q'}_h \leq \sum_{h=i}^{i'-1}t^D_h$.\\
Informally, if the time between $D((x(t),y(t)),d(t))$ entering a region satisfying $\pi$ and then entering a region satisfying $\pi'$ is $\sum_{h=i}^{i'-1}t^D_h$ time units, then the time between $q'(t)$ entering the region satisfying $\pi$ and then entering the region satisfying $\pi'$ is bounded from above by $\sum_{h=i}^{i'-1}t^D_h$. For more intuition about this relations see Fig. 5.
\end{enumerate}
\begin{figure}[htb]
\label{fig:fig3}
\begin{center}
\includegraphics[width=0.5\textwidth]{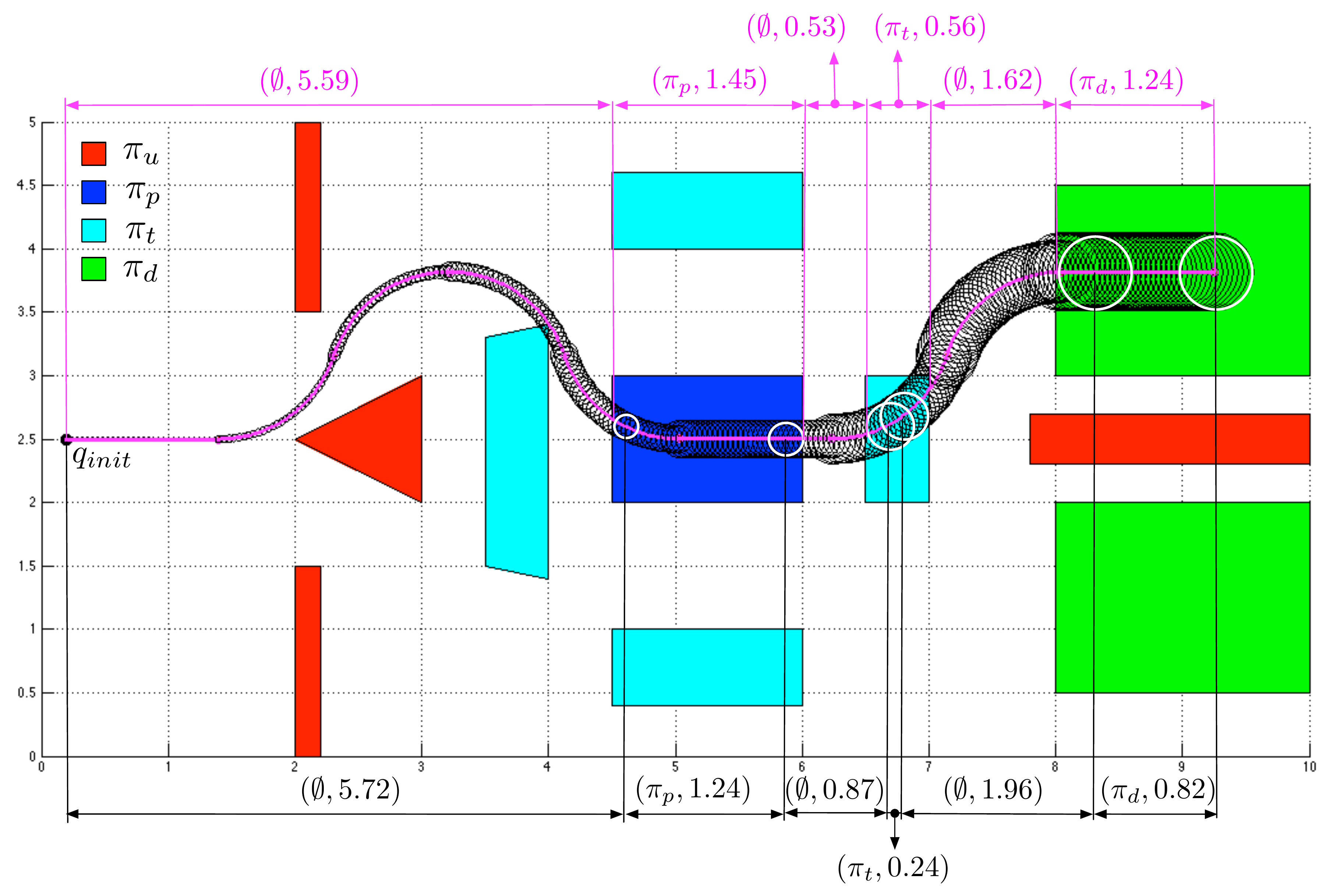}
\caption{An uncertainty region and a sample state (position) trajectory, inside the uncertainty region, are shown in black and magenta, respectively. The corresponding generated traces are $\sigma^D$=$(\emptyset,5.72)(\pi_p,1.24)(\emptyset,0.87)(\pi_t,0.24)(\emptyset,1.96)(\pi_d,0.82)$ and $\sigma^{q'}=(\emptyset,5.59)(\pi_p,1.45)(\emptyset,0.53)(\pi_t,0.56)(\emptyset,1.62)(\pi_d,1.24)$. Let $\phi$  be as given in Example 2. Then, it follows that $\sigma^D \vDash \phi$ and $\sigma^{q'} \vDash \phi$. Note that for $\sigma^{D}_2=\sigma^{q'}_2=\pi_p$, $t^D_2 < t_2^{q'}$ ($1^{\text{st}}$ relation above). Also, for $\sigma^{D}_2=\sigma^{q'}_2=\pi_p$ and $\sigma^{D}_4=\sigma^{q'}_4=\pi_t$,  $\sum_{i=2}^{3}t_i^{q'} < \sum_{i=2}^{3}t_i^D $ ($2^{\text{nd}}$ relation above). }
\end{center}
\end{figure}
Assuming $\sigma^{D} \vDash \phi$, then $\forall j \in \{1,\ldots,f\}$, $\exists i_j,k_j \in \mathbb{N}$ and some $n \in \{1,\ldots,n_j\}$ such that $\sigma^D_{i_j} \vDash \phi_j$ (see Def. \ref{def:def4}). Then, from Prop. \ref{prop:prop2} and Def. \ref{def:def3} and \ref{def:def5}, it follows that $\forall j \in \{1,\ldots, f\}$, $\exists s_j, z_j \in \mathbb{N}$ such that:
\begin{enumerate}
\item  $o_{{s_j}+z_j}^{q'} \in \Pi_j^n$,
\item  for each $s_j \leq i < s_j+z_j$, $o_{i}^{q'} \neq \pi_u$,
\item $\sum_{i=s_j}^{s_j+z_j-1}t_{i}^{q'} \leq \sum_{i=i_j}^{i_j+k_j-1}t_{i}^{D} \leq T_j$ (${2^{\text{nd}}}$ relation above),
\item $t_{{s_j}+z_j}^{q'} \geq t_{i_j+k_j}^D \geq \tau_j^n$ ($1^{\text{st}}$ relation above).
\end{enumerate}
where $s_{j+1}=s_j+z_j$ with $s_1=1$.\\
Thus, $\forall j \in \{1,\ldots,f\}$, $\sigma^{q'}_{s_j} \vDash \phi_j$, and according to Def. \ref{def:def4}, it follows that $\sigma^{q'} \vDash {\phi}$. In Fig. 5 we give an example. $\blacksquare$

\section{Vehicle Control Strategy}
\label{sec:vehicle_control_strategy}
Given the MDP $M$, the next step is to obtain a control policy that maximizes the probability of generating a path through $M$ such that the corresponding trace (as defined in Sec. VI and VII) is satisfying. There are existing approaches that, given an MDP and a temporal logic formula, generate an exact control policy that maximizes the probability of satisfying the specification. In general, exact techniques rely on reasoning about the entire state space, which is a limiting factor in their applicability to large problems. Given $U_r$, $U_l$, $n_r$, $n_l$ and $K$, the size of the MDP $M$ is bounded above by $(|U_r| \times |U_l| \times n_r \times n_l)^K$. Even for a simple case study, due to the size of $M$, using the exact methods to obtain a control policy is computationally too expensive. Therefore, we decide to trade-off correctness for scalability and use computationally efficient techniques based on system sampling.

\subsection{Overview}
We obtain a suboptimal control policy by iterating over the {\emph{control synthesis }} and the {\emph{probability estimation}} procedure until the stopping criterion is met (see Sec. VIII-C). In the control synthesis procedure we use the control synthesis approach from \cite{StatisticalModelCheckingforMDPs} to generate a control policy for the MDP $M$. In particular we use a {\emph{control policy optimization}} part of the algorithm which consists of the {\emph{control policy evaluation}} and the {\emph{control policy improvement}} procedure to incrementally improve a candidate control policy (control policy is initialized with a uniform distribution at each state). Next, in the probability estimation procedure we use SMC by BIE, as presented in \cite{Zuliani:2010:BSM:1755952.1755987}. We estimate the probability that the MDP $M$, under the candidate control policy,  generates a path such that the corresponding trace satisfies BLTL formula $\phi$. Finally, if the estimated probability converges, i.e., if the stopping criterion is met, we map the control policy to a vehicle control strategy. Otherwise, the control synthesis procedure is restarted using the latest update of the control policy. The flow of this approach is depicted in Fig. 6.

\begin{figure}[htb]
\label{fig:fig4}
\begin{center}
\includegraphics[width=0.485\textwidth]{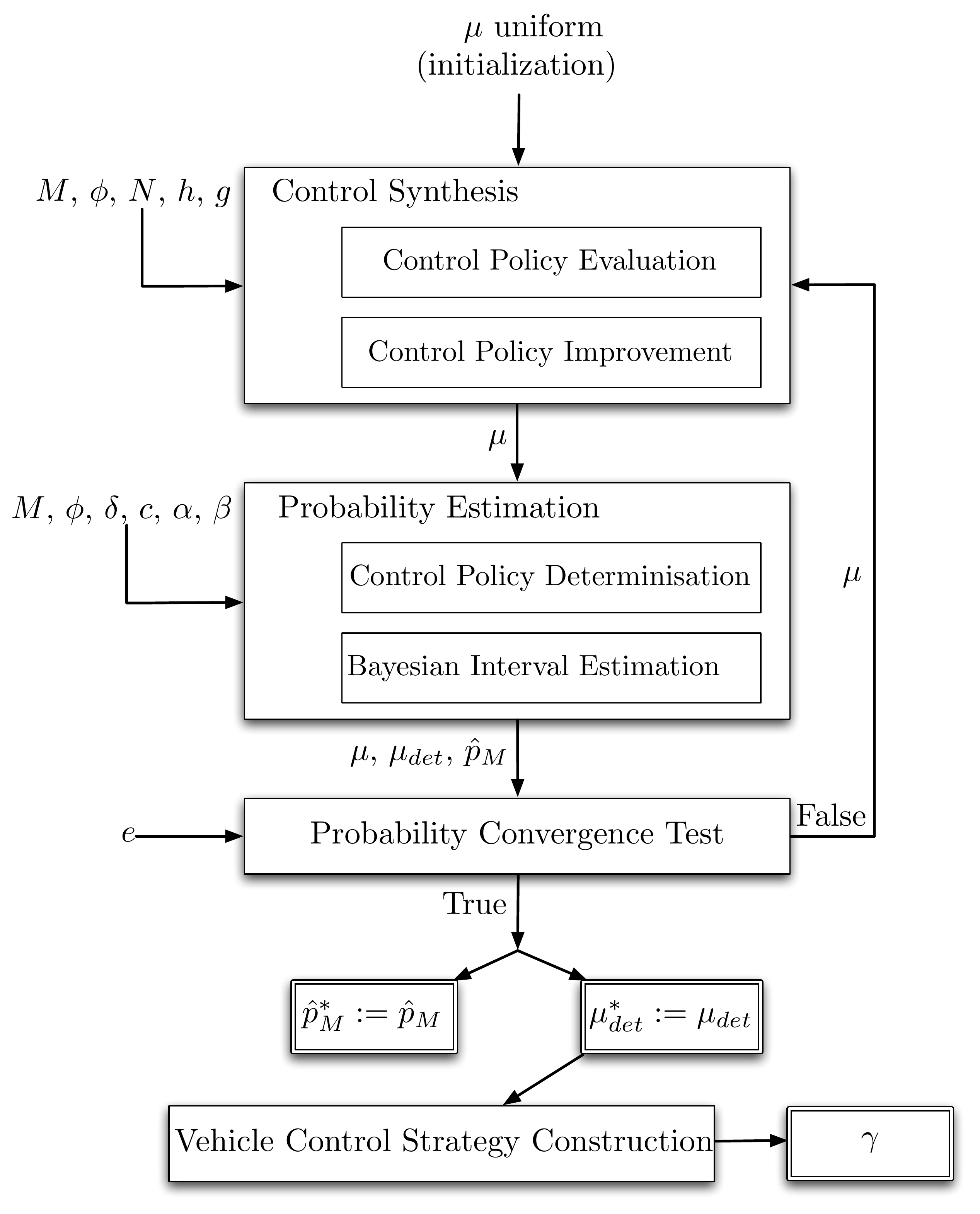}
\caption{Flow chart of the approach used to obtain the vehicle control strategy.}
\end{center}
\end{figure}

\subsection{Control synthesis}
The details of the control policy optimization algorithm can be found in  \cite{StatisticalModelCheckingforMDPs} and here we only give an informal overview of the approach. In the control policy evaluation procedure we sample paths of the MDP $M$ under the current control policy $\mu$. Given a path $\omega=s_0 \xrightarrow{a^1}s_1\xrightarrow{a^2}s_2\ldots s_{K-1}\xrightarrow{a^K}s_K$, where $a^k=(u_r^k,u_l^k)$, the corresponding trace $\sigma$ is generated as described in Sec. VI and VII. Next, we check formula $\phi$ on each $\sigma$ and estimate how likely it is for each action to lead to the satisfaction of BLTL formula $\phi$, i.e., we obtain the estimate of the probability that a path crossing a state-action pair, $(s^k,a^{k+1})$, $k=0,\ldots,K-1$, in $\omega$ will generate a trace that satisfies $\phi$. These estimates are then used in the control policy improvement procedure, in which we update the control policy $\mu$ by reinforcing the actions that led to the satisfaction of $\phi$ most often. The authors (\cite{StatisticalModelCheckingforMDPs}) show that the updated control policy is provably better than the previous one by focusing on more promising regions of the state space. 

The algorithm takes as input MDP $M$, BLTL formula $\phi$ and the current control policy $\mu$, together with the parameters of the algorithm (a greediness parameter $0 < g < 1$, a history parameter $0< h<1$, and the number of sample paths in control policy evaluation procedure, denoted by $N$), and returns the updated probabilistic control policy $\mu$.  In the next step, to estimate the probability of satisfaction, we use the deterministic version of $\mu$, denoted $\mu_{det}$ where: for all $s \in S$ and $a \in A$, $$\mu_{det}(s,a)=I\{{a=\text{arg max}_{a \in Act(s)}\mu(s,a)}\}.$$ In words, we compute a control policy that always picks the best estimated action at each state. 

\subsection{Probability estimation}
Next, we determine the estimate of the probability that the MDP $M$, under the deterministic control policy $\mu_{det}$, generates a path such that the corresponding trace satisfies BLTL formula $\phi$. To do so we use the BIE algorithm as presented in \cite{Zuliani:2010:BSM:1755952.1755987}. We denote the exact probability as $p_M$ and the estimate as $\hat{p}_M$. 

The inputs of the algorithm  are the MDP $M$, control policy $\mu_{det}$, BLTL formula $\phi$, half interval size $\delta \in (0,\frac{1}{2})$, interval coefficient $c \in (\frac{1}{2},1)$, and the coefficients $\alpha, \beta$ of the Beta prior. The algorithm returns $\hat{p}_M$. The algorithm generates traces by sampling paths through $M$ under $\mu_{det}$ (as described in Sec. VI and VII) and checks whether the corresponding traces satisfy $\phi$, until enough statistical evidence has been found to support the claim that $p_M$ is inside the interval $[\hat{p}_M-\delta,\hat{p}_M+\delta]$ with arbitrarily high probability, i.e., $\text{Pr}(p_M \in [\hat{p}_M-\delta,\hat{p}_M+\delta]) \geq c$. 

We stop iterating over the control synthesis and the probability estimation procedure when the difference between the two consecutive probability estimates  converges to a neighborhood of radius $e \in (0,1)$, i.e., when the difference is smaller or equal to $e$. Let $\mu^*_{det}$ and $\hat{p}_M^*$ be the current control policy and the corresponding probability estimate, respectively, when the stopping criterion is met.

\subsection{Control strategy}
The vehicle control strategy is a function $\gamma: S \rightarrow \{U_r \times U_l\}$ that maps a sequence of pairs of measured intervals, i.e., a state of the MDP, to the control inputs:
\begin{equation}
\gamma(({\mathbb{W}^1,\ldots,\mathbb{W}^k}))=\gamma(s_k)=\text{arg max}_{a \in Act(s_k)} \mu^*_{det} (s_k,a),
\end{equation}
$k=1,\ldots,K-1$ with $\gamma(s_0)=\text{arg max}_{a \in Act(s_0)} \mu^*_{det} (s_0,a)$.  

At stage $k$, the control inputs are $$(u_r^k,u_l^k)=\gamma(({\mathbb{W}^1,\ldots,\mathbb{W}^{k-1}})) \in \{U_r \times U_l\}.$$ Thus, given a sequence of pairs of measured intervals, $\gamma$ returns the control inputs for the next stage; the control inputs are equal to the action returned by $\mu^*_{det}$ at the state of the MDP corresponding to that sequence. 

\begin{thm}
The probability that the system given by Eqn. (1), under the vehicle control strategy $\gamma$, generates a state trajectory that satisfies BLTL formula $\phi$ (Eqn. (2)) is bounded from below by $p_M^*$, where $\text{Pr}(p_M^* \in [\hat{p}_M^* - \delta, \hat{p}_M^*+\delta]) \geq c$.
\end{thm}
{\bf{Proof:}} Let $\omega$ be a path through the MDP $M$ and $D((x(t),y(t)),d(t))$ the corresponding uncertainty region as defined in Sec. VI. Let $q'(t)$ be any state trajectory as defined in Prop. 2. Also, let $\sigma^D$ and $\sigma^{q'}$ be the corresponding traces. Trace $\sigma^D$ can (i) satisfy $\phi$ and (ii) not satisfy $\phi$. 

Let us first consider the former. If $\sigma^D \vDash \phi$  from Prop. 3 it follows that $\sigma^{q'} \vDash \phi$. Under $\gamma$ the probability of generating $q'(t)$ is equivalent to generating path $\omega$ under $\mu^*_{det}$. Since under $\mu_{det}^*$ the probability that a path through the MDP $M$ generates a satisfying trace is $p_M^*$ it follows that the probability that the system given by Eqn. (1), under $\gamma$, will generate a satisfying state trajectory is also $p_M^*$.

To show that $p_M^*$ is the lower bound we need to consider the latter case. It is sufficient to observe that because of the conservative approximation of $D((x(t),y(t)),d(t))$ it is possible that $\sigma^{q'}$ satisfies $\phi$, even though $\sigma^D$ does not satisfy it. Therefore, it follows that the probability that system given by Eqn. (1), under the vehicle control strategy $\gamma$, generates a state trajectory that satisfies BLTL formula $\phi$, is bounded from below by $p_M^*$. The rest of the proof, i.e., $\text{Pr}(p_M^* \in [\hat{p}_M^* - \delta, \hat{p}_M^*+\delta]) \geq c$, is given in  \cite{Zuliani:2010:BSM:1755952.1755987}. $\blacksquare$

\subsection{Complexity}
As stated above, the size of the MDP $M$ is bounded above by $(|U_r| \times |U_l| \times n_r \times n_l)^K$. 
Obviously, it can be expensive (in sense of memory usage) to store the whole MDP. Since our approach is sample-based, it is not necessary for the MDP to be constructed explicitly. Instead, a state of the MDP is stored only if it is sampled during the control synthesis procedure. As a result, during the execution, the number of states stored in the memory is bounded above by $N \times K \times n$, where $n$ is the number of iterations between the control synthesis and the probability estimation procedures.

The complexity analysis of the control synthesis part can be found in  \cite{StatisticalModelCheckingforMDPs} and the complexity analysis of BIE algorithm can be found in \cite{Zuliani:2010:BSM:1755952.1755987}.

\section{Case study}
\label{sec:case_study}
We considered the system given by Eqn. (1) and we used the numerical values corresponding to Dr. Robot's x80Pro mobile robot equipped with two incremental encoders. The parameters were $r=0.085$m  and $L=0.295$m. To reduce the complexity, $\{U_r \times U_l\}$  was limited to $\{(\frac{1 + L}{4r},\frac{1- L}{4r}), (\frac{1}{4r},\frac{1}{4r}),(\frac{1- L}{4r},\frac{1+ L}{4r})\}$, where the pairs of control inputs corresponded to a vehicle turning left at $\frac{1}{2}$$\frac{\text{rad}}{\text{s}} $, going straight, and turning right at $\frac{1}{2}$$\frac{\text{rad}}{\text{s}}$, respectively, when the forward speed is $\frac{1}{4}$$\frac{\text{m}}{\text{s}}$.

{\emph{Measurement resolution:}} To obtain the angular wheel velocity, the frequency counting method \cite{4510607} was used, i.e., the encoder pulses inside a given sampling period were counted. The number of pulses per revolution (i.e., the number of windows in the code track of the encoders) was $378$ and the sampling period was set to $\Delta t=2.6$s. Thus, according to \cite{4510607} the measurement resolution was $\Delta \epsilon_r = \Delta \epsilon_l = \frac{2 \pi}{378 \cdot 2.6} \approx 0.0064$.

{\emph{Probability density functions:}} 
We obtained the distributions through experimental trials. Specifically, we used control inputs from $\{U_r \times U_l\}$ as the robot inputs and then measured the actual angular wheel velocities using the encoders. We obtained $\epsilon_i^{min}$ ($\epsilon_i^{max}$) by taking the minimum (maximum) over $\{\underline{\epsilon}_i^1,\ldots,\underline{\epsilon}_i^{k}\}$ ($\{\overline{\epsilon}_i^1,\ldots,\overline{\epsilon}_i^{k}\}$), where $[\underline{\epsilon}_i^j,\overline{\epsilon}_i^j]$, $j \in \{1,\ldots,k\}$, $i \in \{r,l\}$, was the noise interval, of length $\Delta \epsilon_i$, determined from the $j$-th measurement of the encoder $i$ and $k$ was the total number of measurements. Note that $n_i=\frac{|\epsilon_i^{max}-\epsilon_i^{min}|}{\Delta \epsilon_i}$, $i \in \{r,l\}$.  Finally, the probabilities for Eq. (4) that defined the probability density functions, were equal to the number of times a particular noise interval was measured over $k$. For $k=150$ (i.e., by using each control input from  $\{U_r \times U_l\}$ $50$ times) we obtained $-\epsilon_r^{min}=\epsilon_r^{max}=-\epsilon_l^{min}=\epsilon_l^{max}=0.0096$ and the corresponding probabilities. 

The set of propositions was $\Pi=\{\pi_u,\pi_p,\pi_{t1}, \pi_{t2},\pi_d\}$ where $\pi_u,\pi_p,\pi_{t1},\pi_{t2},\pi_d$ labeled the {\ttfamily{unsafe}}, {\ttfamily{pick-up}}, {\ttfamily{test1}}, {\ttfamily{test2}} and the {\ttfamily{drop-off}} regions, respectively. The motion specification was: 

{\emph{
Start from an initial state $q_{init}$ and reach a {\ttfamily{pick-up}} region within $14$ time units and stay in it at least $0.8$ time units, to pick-up the load. After entering the {\ttfamily{pick-up}} region,  reach a {\ttfamily{test1}} region within $5$ time units and stay in it at least $1$ time units or reach a {\ttfamily{test2}} region within $5$ time units and stay in it at least $0.8$ time units. Finally, after entering the {\ttfamily{test1}} region or the {\ttfamily{test2}} region reach a {\ttfamily{drop-off}} region within $4$ time units to drop off the load. Always avoid the {\ttfamily{unsafe}} regions. 
}}

The specification translates to BLTL formula $\phi$:
\begin{equation}
\begin{split}
\label{eq:eq10}
\phi=\neg \pi_u  \mathbf{U}^{\leq 14}(\mathbf{G}^{\leq 0.8}\pi_p \wedge \neg \pi_u \mathbf{U}^{\leq 5}(\\
[\mathbf{G}^{\leq 1}\pi_{t1} \vee \mathbf{G}^{\leq 0.8}\pi_{t2}] \wedge \neg \pi_u \mathbf{U}^{\leq 4} \pi_d)).
\end{split}
\end{equation}

Two different environments are shown in Fig. 7. The estimated probability $\hat{p}_M^*$ corresponding to environment $A$ and $B$ was $0.664$ and $0.719$, respectively. From Eq. (\ref{eq:eq10}) it followed that $K=9$. The numerical values in the control synthesis procedure and the probability estimation procedure were as follows: $N=10000$, $h=0.6$, $g=0.6$, $\delta = 0.05$, $c=0.95$, $\alpha=\beta=1$, and $e=0.05$. For both environments, we found the vehicle control strategy through the method described in Sec. VIII. 

\begin{figure}[htb]
\label{fig:fig5}
\begin{center}
\includegraphics[width=0.49\textwidth]{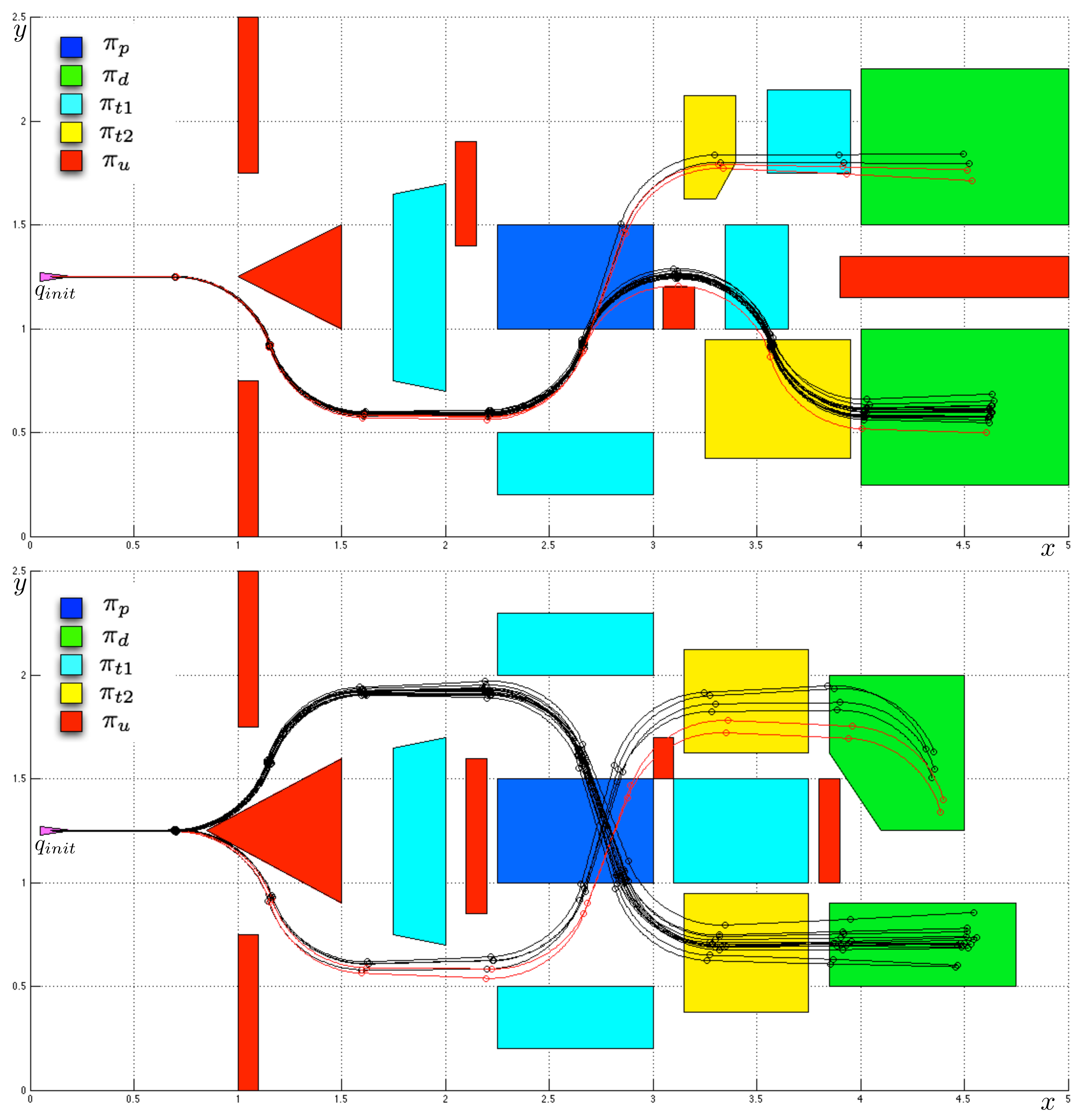}
\caption{20 sample state (position) trajectories for cases $A$ and $B$ (to be read top-to-bottom). The {\ttfamily{unsafe}}, {\ttfamily{pick-up}}, {\ttfamily{test1}}, {\ttfamily{test2}}, and the {\ttfamily{drop-off}} regions are shown in red, blue, cyan, yellow and green, respectively. Satisfying and violating trajectories are shown in black and red, respectively. Note that, in case $A$, the upper two red trajectories avoid the {\ttfamily{unsafe}} regions and visit the {\ttfamily{pick-up}}, {\ttfamily{test2}}, and the {\ttfamily{drop-off}} region in the correct order, but they violate the specification because they do not stay long enough in the {\ttfamily{test2}} region. }
\end{center}
\end{figure}

Since it is not possible to obtain the exact probability that the system given by Eqn. (1), under the vehicle control strategy, generates a satisfying state trajectory, in order  to verify our result (Theorem 1), we  performed multiple runs of BIE algorithm by simulating the system under the vehicle control strategy (using the same numerical values as stated above and by generating traces as described in Sec. IV). We denote the resulting probability estimate as $\hat{p}_S$ and we compare it to $\hat{p}_M^*$.

\begin{table}[h]
\caption{Probability estimates of satisfying the specification}
\label{table:data}
\begin{center}
\scalebox{1}{
\begin{tabular}{| c || c || c || c || c |}
\hline 
Environment & $\hat{p}_M^*$ & \multicolumn{3}{c|}{$\hat{p}_S$}\\ 
\cline {3-5}&  &  Run 1 &  Run 2 & Run 3\\
\hline
$A$ & $0.664$ & $0.847$ & $0.832$ & $0.826$\\
\hline
$B$ & $0.719$ &  $0.891$ & $0.898$ & $0.879$ \\
\hline
\end{tabular}
}
\end{center}
\end{table}

In Fig. 7 we show sample state trajectories and in Table \ref{table:data} we compare the estimated probabilities obtained on the MDP, $\hat{p}^*_M$, with the estimated probabilities obtained by simulating the system, $\hat{p}_S$. The results support Theorem 1, since $\hat{p}_S$ is bounded from below by $\hat{p}^*_M$. The discrepancy in the probabilities is mostly due to the conservative approximation of the uncertainty region in Sec. VI. The Matlab code used to obtain the vehicle control strategy ran for approximately 2.2 hours on a computer with a 2.5GHz dual processor. 

In Fig. 8 we show a sample run of the robot in environment $A$. A projector was used to display the environment and the state (position) trajectory was reconstructed using the OptiTrack (\href{http://www.naturalpoint.com/optitrack}{http://www.naturalpoint.com/optitrack}) system with eight cameras.
\begin{figure}[htb]
\label{fig:fig5}
\begin{center}
\includegraphics[width=0.48\textwidth]{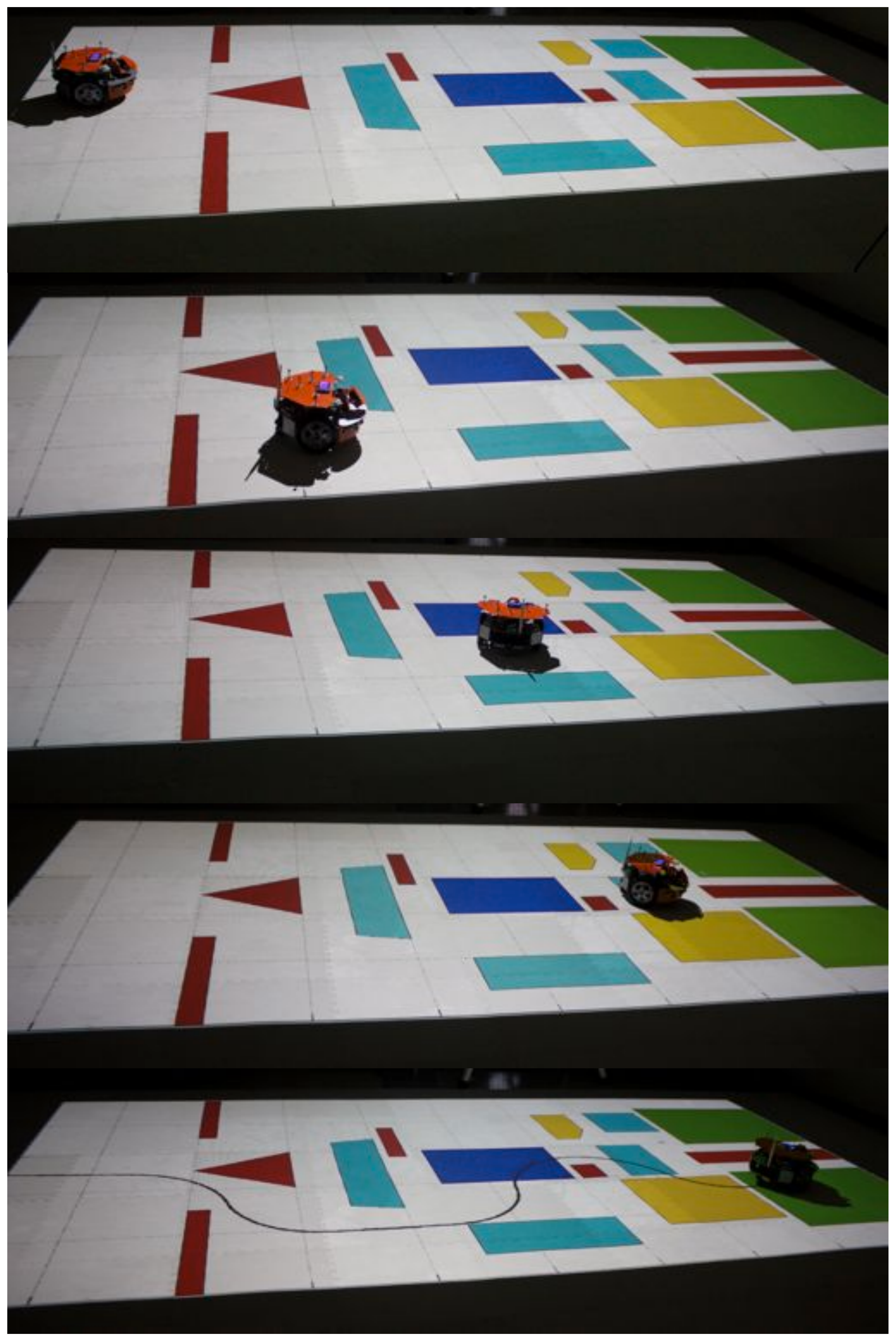}
\caption{Snapshots (to be read top-to-bottom) from a movie (available online at \urlwofont{http://people.bu.edu/icizelj/Igor_Cizelj/diff-bltl.html}) showing a robot motion produced by applying the vehicle control strategy for environment $A$. The generated trajectory satisfied $\phi$ (Eq. (\ref{eq:eq10})).}
\end{center}
\end{figure}

\section{Discussion}
\label{sec:conclusion}
We developed a feedback control strategy for a stochastic differential drive mobile robot such that the probability of satisfying a time constrained specification given in terms of a temporal logic statement is maximized. By mapping sensor measurements to a Markov Decision Process (MDP) we translate the problem to finding a control policy maximizing the probability of satisfying a Bounded Linear Temporal Logic (BLTL) formula on the MDP. The solution is based on Statistical Model Checking for MDPs and we show that the probability that the vehicle satisfies the specification is bounded from below by the probability of satisfying the specification on the MDP. 

The key limitation of the proposed approach is the computation time.  Since our algorithm is based on Statistical Model Checking for MDPs presented in \cite{StatisticalModelCheckingforMDPs}, to put the running time of our algorithm into perspective, we compare it to the running time of  Statistical Model Checking for MDPs when dealing with the following motion planning study: 
each of the  two robots living in a $20 \times 20$ grid world must pick up some object and then meet with the other robot within a certain time bound, while avoiding unsafe grids. At each point in time, either robot can try to move 10 grid units in any of the four cardinal directions, but each time a robot moves, it has some probability of ending up somewhere in a radius of $3$ grid units of the intended destination. Statistical Model Checking for MDPs (when $c=0.95$) solves this problem in approximately $20$ minutes. Now, let us consider the algorithm and the case study presented in this paper. 
First, note that in order  for Theorem 1 to hold, when generating a trace corresponding to an uncertainty region, we have to perform a computationally expensive step of taking the intersection between the uncertainty region and all of the regions of interest (see Def. 5, Sec. VII). 
Second, note that we are dealing with a system that is continuous both in space and time. Therefore, at each time step, when constructing an uncertainty region, the algorithm is required to perform multiple integrations of the system. Thus, the increase in the computational complexity, in order to have probabilistic guarantees for the original system, is the reason the running time of our algorithm (when $c=0.95$) is approximately $2.2$ hours. 

Since sampling (i.e., generating traces) accounts for the majority of our runtime, future work includes improving the sampling performance and making the implementation fully parallel. 
Additionally, to address the problem of discrepancy between the probabilities obtained on the MDP and the probabilities obtained by simulating the system the future work also includes developing a less conservative uncertainty model. 

\section{Acknowledgements}
The authors gratefully acknowledge David Henriques from Carnegie Mellon University  for comments on an earlier draft. Also, the authors would like to thank Benjamin Troxler, Michael Marrazzo and Matt Buckley from Boston University for their help with the experiments.  

\bibliographystyle{alpha} 
\bibliography{references}   
\end{document}